\documentclass[lettersize,journal]{IEEEtran}
\usepackage{amsmath,amsfonts}
\usepackage{algorithmic}
\usepackage{algorithm}
\usepackage{array}
\usepackage[caption=false,font=normalsize,labelfont=sf,textfont=sf]{subfig}
\usepackage{textcomp}
\usepackage{stfloats}
\usepackage{url}
\usepackage{verbatim}
\usepackage{graphicx}
\usepackage{cite}
\usepackage[colorlinks,
            linkcolor=blue,
            anchorcolor=blue, 
            urlcolor=blue,
            citecolor=blue]{hyperref}
\usepackage{booktabs}
\usepackage{multirow}
 \usepackage{subfig}
 \usepackage{caption}

\makeatletter
\renewcommand{\maketag@@@}[1]{\hbox{\m@th\normalsize\normalfont#1}}%
\makeatother

\begin{document}

\title{Empathy Level Alignment via Reinforcement Learning for Empathetic Response Generation}

\author{Hui Ma, Bo Zhang, Bo Xu, Jian Wang, Hongfei Lin, and Xiao Sun
\thanks{Hui Ma (Corresponding author) and Xiao Sun are with the School of Computer Science and Information Engineering, Hefei University of Technology, Hefei 230601, China (e-mail: huima@hfut.edu.cn; sunx@hfut.edu.cn). \par
Bo Zhang, Bo Xu, Jian Wang, and Hongfei Lin are with the School of Computer Science and Technology, Dalian University of Technology, Dalian 116024, China (e-mail: zhangbo1998@mail.dlut.edu.cn; xubo@dlut.edu.cn;
wangjian@dlut.edu.cn; hflin@dlut.edu.cn).
}}
%


\markboth{Journal of \LaTeX\ Class Files,~Vol.~14, No.~8, August~2021}%
{Shell \MakeLowercase{\textit{et al.}}: A Sample Article Using IEEEtran.cls for IEEE Journals}


\maketitle

\begin{abstract}
Empathetic response generation, aiming to understand the user's situation and feelings and respond empathically, is crucial in building human-like dialogue systems. Traditional approaches typically employ maximum likelihood estimation as the optimization objective during training, yet fail to align the empathy levels between generated and target responses. To this end, we propose an empathetic response generation framework using reinforcement learning (EmpRL). The framework develops an effective empathy reward function and generates empathetic responses by maximizing the expected reward through reinforcement learning. EmpRL utilizes the pre-trained T5 model as the generator and further fine-tunes it to initialize the policy. To align the empathy levels between generated and target responses within a given context, an empathy reward function containing three empathy communication mechanisms---emotional reaction, interpretation, and exploration---is constructed using pre-designed and pre-trained empathy identifiers. During reinforcement learning training, the proximal policy optimization algorithm is used to fine-tune the policy, enabling the generation of empathetic responses. Both automatic and human evaluations demonstrate that the proposed EmpRL framework significantly improves the quality of generated responses, enhances the similarity in empathy levels between generated and target responses, and produces empathetic responses covering both affective and cognitive aspects.
\end{abstract}


\section{Introduction}
\IEEEPARstart{O}{pen}-domain dialogue systems, which converse with humans in open domains to provide reasonable responses and entertainment\cite{chen2017survey}, have been extensively studied in recent years. With the rapid development of neural response generation models\cite{gu2021dialogbert,li2022hier,zhao2022towards,belainine2023end,feng2023less}, the fluency and coherence of responses have significantly improved. Despite these advancements, the lack of empathy still results in a noticeable gap between such systems and truly human-like dialogue systems.\par
\begin{figure}[!t]
\hspace{-2mm}
\subfloat[\scriptsize Empathetic dialogue responses]{\label{fig:0a}
\includegraphics[width=1.72in]{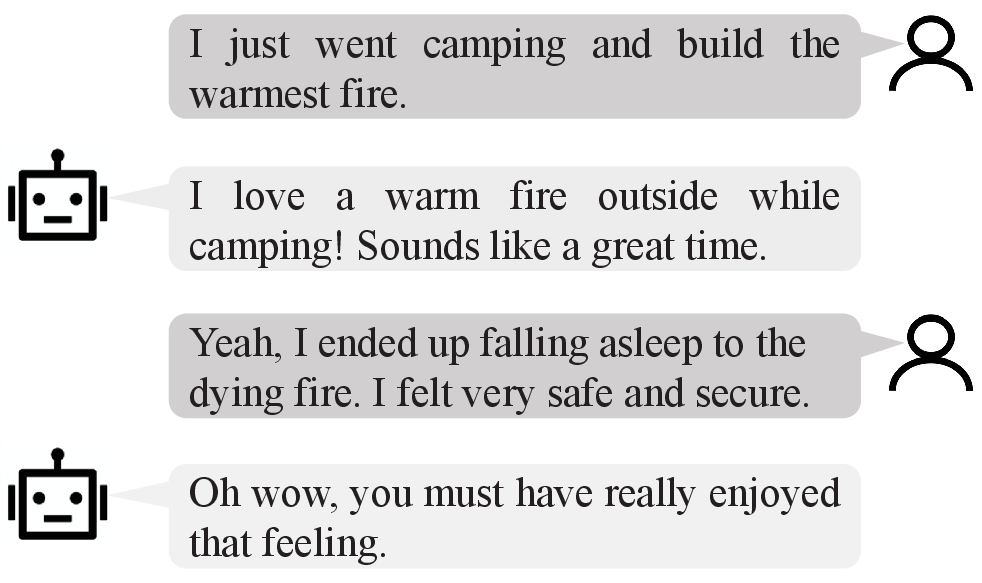}}
 \hspace{-2mm}
\subfloat[\scriptsize Non-empathetic dialogue responses]{\label{fig:0b}
\includegraphics[width=1.72in]{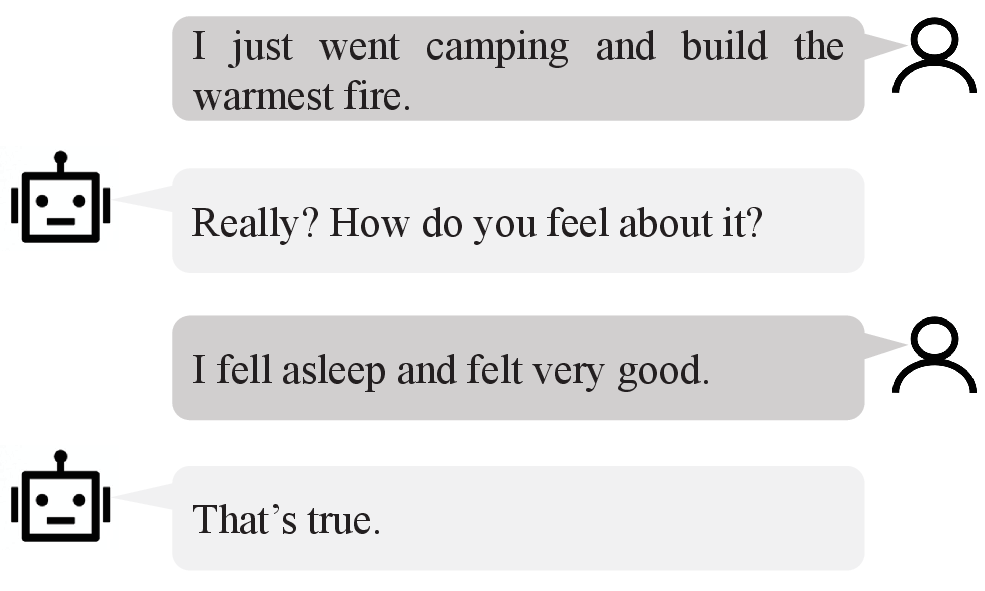}}
\hspace{-5mm}
\caption{An example of two types of dialogue responses.}
\label{fig:0}
\end{figure}

Empathy generally refers to the capacity to understand other's experiences and feelings, encompassing both affective and cognitive aspects\cite{davis1983measuring,cuff2016empathy}. The affective aspect involves emotional simulation in reaction to the user's experiences, while the cognitive aspect focuses on understanding the user's situation. Empathy is a fundamental characteristic of human communication, and exploring ways to generate empathetic responses is  crucial for the development of human-like dialogue systems\cite{raamkumar2022empathetic}. It enhances user experience and satisfaction, facilitates deeper human-machine interactions, and provides valuable support in various scenarios, such as psychological counseling, emotional companionship, and mental health support. Fig.~\ref{fig:0} illustrates examples of both empathetic and non-empathetic dialogue responses.\par
Rashkin et al.\cite{rashkin2019towards} released a large-scale empathetic dialogue dataset EmpatheticDialogues, paving the way for research in empathetic response generation. Existing methods are primarily divided into two categories: one focuses on affective empathy, detecting and utilizing the user's emotion to generate responses, such as MoEL\cite{lin2019moel}, MIME\cite{majumder2020mime}, and KEMP\cite{li2022knowledge}; the other considers both affective and cognitive aspects, such as CoMAE\cite{zheng2021comae}, CEM\cite{sabour2022cem}, and CASE\cite{zhou2022case}. These two categories mainly rely on supervised learning, using maximum likelihood estimation as the optimization objective during training. However, these models overlook the alignment of empathy levels between the generated responses and target responses.\par
\textit{Empathy level} is a fundamental concept in empathy theory, introduced by Sharma et al.\cite{sharma2020computational}. They pointed out that empathy expression is manifested through three key communication mechanisms: emotional reaction, interpretation, and exploration. Emotional reaction represents affective empathy, while interpretation and exploration reflect cognitive empathy. Each mechanism is categorized into no, weak, or strong communication to quantify the empathy levels of responses within a given context. Aligning the empathy levels between generated and target responses facilitates a closer approximation of human empathy expression, thereby enhancing the quality of generated empathetic responses.\par
To this end, we propose an empathetic response generation framework using reinforcement learning (EmpRL). The framework develops an effective empathy reward and maximizes the expected reward through reinforcement learning (RL) to generate empathetic responses. EmpRL uses the pre-trained T5 model as 
the generator, fine-tunes it to generate fluent responses, and utilizes the fine-tuned generator to initialize the policy. Subsequently, an empathy identifier is designed to recognize the empathy levels of responses within the dialogue context and is trained on the Mental Health Subreddits dataset. The pre-trained empathy identifier is then used to construct an empathy reward function, which incorporates three empathy communication mechanisms, ensuring that empathy levels of the generated responses align more effectively with the target responses. Finally, the proximal policy optimization (PPO) algorithm is employed to train the policy for generating empathetic responses. We conduct extensive experiments on the benchmark dataset EmpatheticDialogues to evaluate the performance of our proposed EmpRL framework. Both automatic and human evaluation results demonstrate that EmpRL enhances the similarity in empathy levels between generated responses and target responses, and improve the quality of the generated responses.\par
Our contributions are summarized as follows:
\begin{itemize}
\item We propose EmpRL, an RL-based framework for empathetic response generation, which develops an empathy reward function to align the empathy levels between generated and target responses.
\item The experimental results demonstrate that our proposed EmpRL exhibits superior performance and generates more empathetic responses encompassing both affective and cognitive dimensions.
\end{itemize}
The remainder of this paper is organized as follows: Section~\ref{sec:related_work} reviews related research on empathetic response generation and RL-based sequence generation. Section~\ref{sec:method} defines the task and introduces the proposed EmpRL framework. Experimental settings and results are presented in Section~\ref{sec:experiment_settings} and \ref{sec:results}, respectively. Finally, Section~\ref{sec:conclustion} concludes the paper, and Section~\ref{sec:futurework} outlines directions for future work.
\section{Related Work}
\label{sec:related_work}
\subsection{Empathetic Response Generation}
In recent years, empathetic response generation has gained significant attention. Unlike emotional response generation, which focuses on producing responses with specific emotions\cite{zhou2018emotional,colombo2019affect,shen2020cdl,zhang2023dual}, the task of empathetic response generation aims to respond in an empathetic manner. Existing works are mainly divided into two categories. \par
The former involves detecting and utilizing the user's emotion. MoEL\cite{lin2019moel} employs different Transformer decoders to compute response representations for each possible emotion, which are then softly combined to generate the final response. Majumder et al.\cite{majumder2020mime} proposed that empathetic responses often mimic the user's emotions to some degree, and introduced MIME to generate both mimicking and non-mimicking representations. EmpDG\cite{li2020empdg} leverages coarse-grained dialogue-level and fine-grained token-level emotions to generate empathetic responses, while introducing interaction discriminators to interact with user feedback. KEMP\cite{li2022knowledge} integrates commonsense knowledge and emotional lexical knowledge to understand and express emotions. The latter considers both affective and cognitive empathy. Buliding upon GPT-2\cite{radford2019language}, CoMAE\cite{zheng2021comae} hierarchically models three factors of empathy expression: communication mechanism, dialogue act, and emotion. CEM\cite{sabour2022cem} leverages external commonsense knowledge to acquire comprehensive information about the user's situation and feelings. CASE\cite{zhou2022case} integrates two heterogeneous graphs involving commonsense and concept knowledge, and introduces a two-level strategy to align cognition and affection.\par
Due to the remarkable capabilities of large language models (LLMs) in natural language generation tasks, several studies have explored the potential of LLMs for generating empathetic responses. Loh et al.\cite{loh2023harnessing} proposed a simple instructional prompt to investigate the ability of five LLMs to generate empathetic responses. Qian et al.\cite{qian2023harnessing} conducted an empirical investigation into the performance of LLMs and proposed three targeted methods for improvement. Chen et al.\cite{chen2023soulchat} developed a multi-turn empathetic conversation dataset and fine-tuned ChatGLM to improve its ability to generate empathetic responses.\par
However, existing approaches fail to explicitly align the empathy levels between generated responses and target responses. To ensure that the empathy levels of the generated responses align more closely with the target responses, we develop an effective empathy reward function and use RL training to generate empathetic responses.
\subsection{Reinforcement Learning for Sequence Generation}
RL\cite{sutton2018reinforcement} trains the agent using rewards derived from interactions with the environment, primarily to solve sequential decision-making problems. With the advancement of deep RL, sequence generation using RL has received widespread attention. To alleviate the exposure bias problem in generative models, Ranzato et al.\cite{ranzato2016sequence} proposed mixed incremental cross-entropy reinforce (MIXER), which leverages RL to optimize evaluation metrics such as BLEU and ROUGE. Rennie et al.\cite{rennie2017self} introduced self-critical sequence training (SCST), utilizing CIDEr metric as the reward function, which outperforms MIXER. Ziegler et al.\cite{ziegler2019fine} asked human annotators to provide feedback on various answers generated by the language model to train a reward model. Le et al.\cite{le2022coderl} introduced CodeRL framework for program synthesis tasks, leveraging RL to enhance the code generation capabilities of pre-trained language models.\par
In the field of dialogue generation, Li et al.\cite{li2016deep} devised three distinct reward functions to evaluate the informativity, coherence, and ease of answering of generated responses, and employed the policy gradient method to enhance sequence-to-sequence models. Zhang et al.\cite{zhang2018reinforcing} introduced three coherence models to compute the similarity between the dialogue context and the generated response, and incorporated them as reward functions. Saleh et al.\cite{saleh2020hierarchical} utilized hierarchical RL to model utterance-level rewards, improving the flexibility of dialogue models in learning long-term conversational rewards. Shin et al.\cite{shin2020generating} emphasized the importance of sentiment look-ahead and designed three distinct sentiment look-ahead reward functions to encourage more empathetic responses. Su et al.\cite{ su2024rlca} integrated both cognitive and affective aspects of empathy using RL to enhance the perceptual and emotional expression abilities.\par
Achieving human-like expressions of empathy is a crucial factor in empathetic response generation. In this work, we use RL to align the empathy levels between generated and target responses within a given context, enhancing the quality of generated responses to encompass both affective and cognitive empathy.
\section{Methodology}
\label{sec:method}
\subsection{Task Definition}
The task of empathetic response generation aims to develop a dialogue model that acts as a listener and generates empathetic responses. 
Formally, let ${\cal C}=\left[{{u_1},\,{u_2},\,\cdots,\,{u_{N-1}}} \right]$ denote a dialogue context consisting of $N-1$ utterances, where the $i$-th utterance $u_i=\left[w_1^i,\, w_2^i,\,\cdots, \, w_{M_i}^i \right]$ comprises $M_i$ words. Odd-numbered utterances $\left( {{u_1},{u_3},\cdots,{u_{N-1}}} \right)$ belong to the speaker, while even-numbered utterances $\left( {{u_2},{u_4},\cdots,{u_{N-2}}} \right)$ correspond to the listener. The goal is to generate the next response $y = \left[ {{y_1},\,{y_2},\, \cdots,\,{y_T}} \right]$, which is fluent, coherent with the context, and empathetic to the speaker's situation and feelings. 
\subsection{Overview}

\begin{figure*}
    \centering
\includegraphics[width=4.9in]{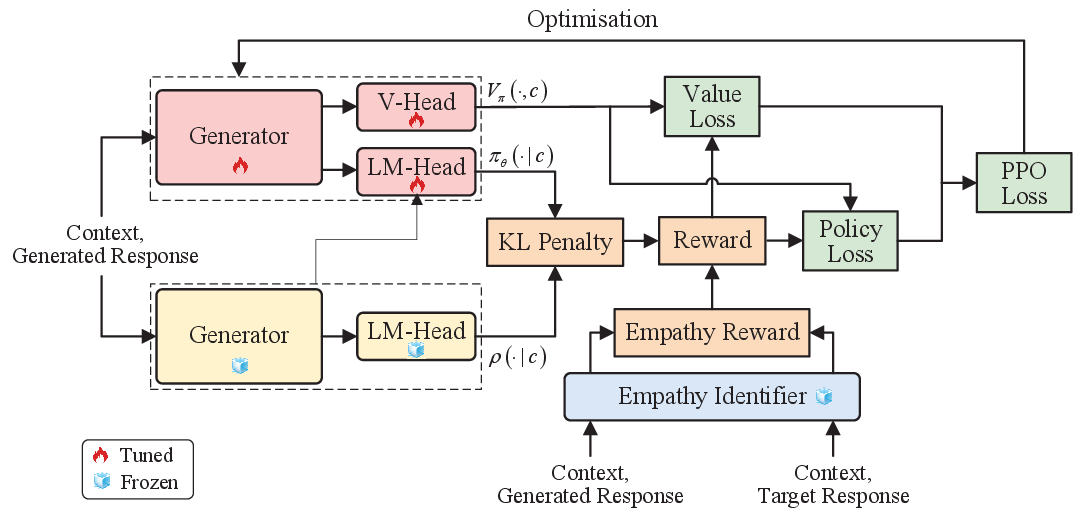}
    \caption{Overall architecture of the proposed EmpRL.}
    \label{fig:emprl}
\end{figure*}
The architecture of our proposed EmpRL framework is illustrated in Fig.~\ref{fig:emprl}. EmpRL utilizes the pre-trained T5 model as a generator, which is fine-tuned to initialize the policy. Subsequently, an empathy reward function is designed, incorporating three empathy communication mechanisms, namely emotional reaction, interpretation, and exploration.
This function aligns the empathy levels between the generated responses and the target responses within a given context. To prevent the policy from deviating excessively from the fine-tuned generator, which could lead to incoherent responses, a KL penalty term is integrated into the empathy reward. Finally, the PPO algorithm is employed to further train the policy, enabling it to generate empathetic responses that encompass both affective and cognitive aspects.\par
In the following sections, we first introduce the generator for policy initialization and the empathy identifier for calculating the empathy reward. Then, we provide a detailed description of the various components of EmpRL, including the state, action, policy, reward function, advantage function, objective, and optimization method.
\subsection{Generator}
The pre-trained T5 model\cite{raffel2020exploring} serves as the backbone for generating responses. We further train the model using full fine-tuning, and the fine-tuned generator is described as follows:
\begin{equation}
	\rho \left( \hat y|c \right) = \prod\limits_{j = 1}^M {\rho \left( {\hat {y}_j|\hat {y}_{ < j},c} \right)}.
\end{equation}
where $c$ represents the context and $\hat y$ is the generated response. 
\subsection{Empathy Identifier}
\begin{figure}[!t]
	\centering
	\includegraphics[width=1.6in]{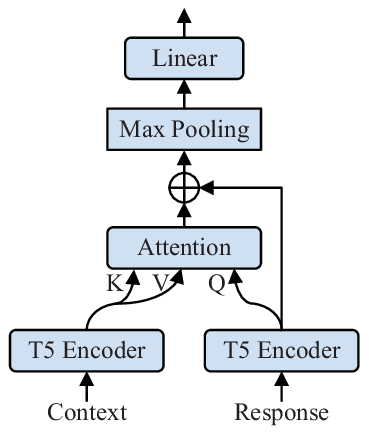}
	\caption{The architecture of Empathy Identifier.}
	\label{fig:emp_iden}
\end{figure}
To establish an effective empathy reward in EmpRL, we propose an empathy identifier that
identifies the empathy level of a response within its context. Fig.~\ref{fig:emp_iden} gives the structure of the empathy identifier. The identifier first utilizes two independently pre-trained T5 encoders\cite{raffel2020exploring} to encode the context and the response, respectively. Then, the encoded response, serving as the query, and the encoded context, serving as the key and value, are input into a single-head attention mechanism\cite{Vaswani2017attn}, followed by a residual connection\cite{he2016deep}, to generate a context-aware representation of the response. Finally, the representation is fed into a max-pooling operation and a linear layer to produce the predicted label, i.e., empathy level.\par
To train the empathy identifier, we use a publicly available dataset---the Mental Health Subreddits dataset\cite{sharma2020computational}. The dataset\footnote{\url{https://github.com/behavioral-data/Empathy-Mental-Health}} is derived from threads posted on $55$ subreddits that focus on mental health, containing 3k \texttt{<seek post, response post>} pairs. For each pair, three empathy communication mechanisms (i.e., emotional reaction, interpretation, and exploration) are individually labeled as no, weak, or strong, indicating the empathy level. The annotators are crowdworkers trained using the EPITOME framework\cite{sharma2020computational} to ensure high-quality annotations. The statistics of the dataset are presented in Table~\ref{tab:statistics}.\par
\begin{figure}[!t]
	\centering
	\includegraphics[width=3.0in]{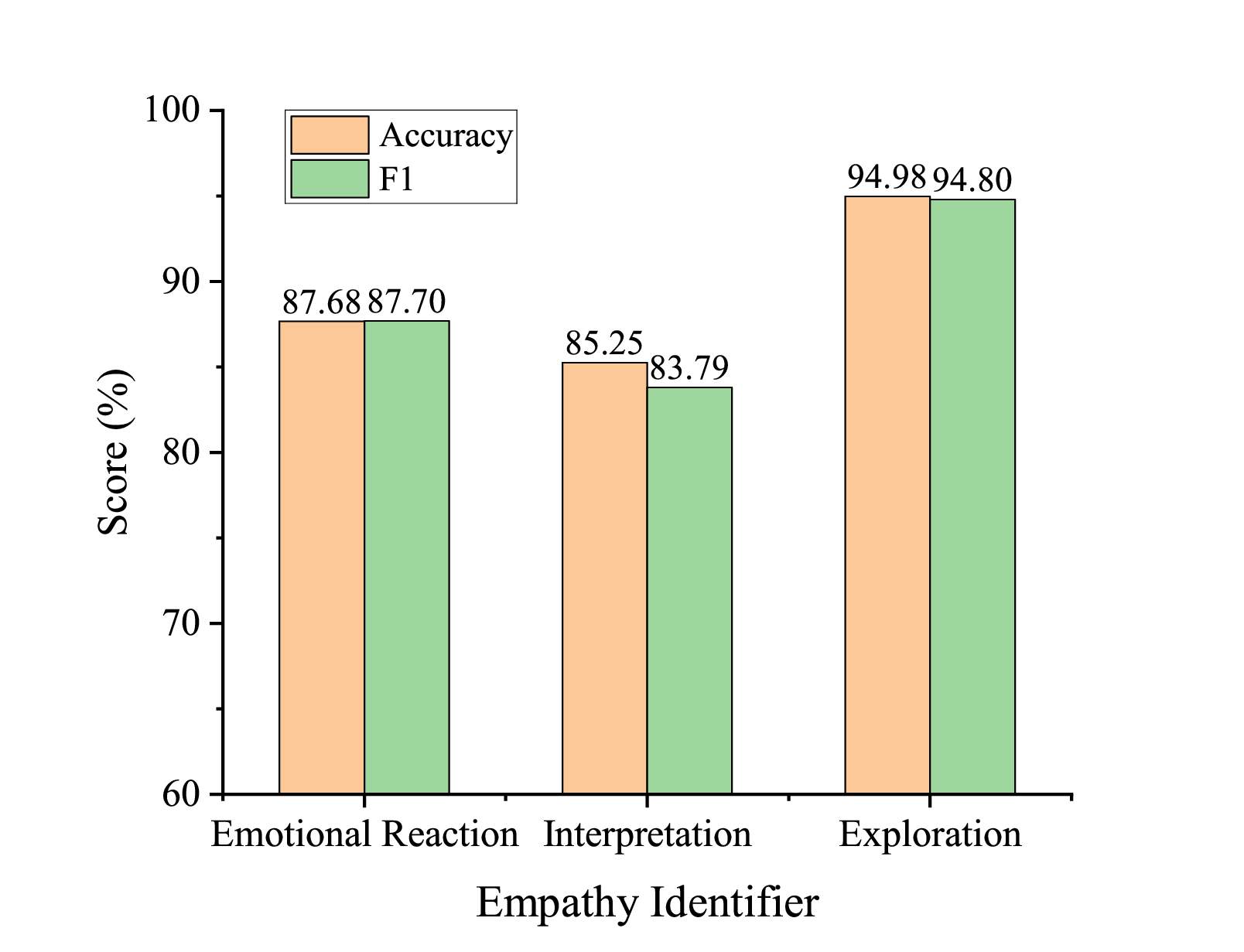}
	\caption{Results of empathy identifiers on Mental Health Subreddits validation set.}
	\label{fig:identifier}
\end{figure}

\begin{table}[!t]
	\centering
	\caption{The statistics of Mental Health Subreddits dataset.}
	\renewcommand\arraystretch{1.25}
	\begin{tabular}{lcccc}
		\toprule
		Communication Mechanisms          & No & Weak    & Strong & Total \\ 
		\midrule
		Emotional Reaction & 2$,$037 & 895 & 152 &  \multirow{3}{*}{3$,$084 }\\
		Interpretation  &  2$,$604 &  104 & 376  \\
		Exploration   & 1$,$626 & 114 &   1$,$344   \\ 
		\bottomrule
	\end{tabular}
	\label{tab:statistics}
\end{table}
Three empathy identifiers are trained, each tailored to a specific communication mechanism. We set aside 20\% of the dataset for validation, and Fig.~\ref{fig:identifier} shows the validation results. We find that all three identifiers achieve excellent performance, demonstrating the feasibility of the proposed empathy identifier architecture.
\subsection{EmpRL Framework}
The proposed EmpRL framework treats empathetic response generation as a sequential decision-making problem and leverages RL to maximize the expected reward for generating responses. Below, we provide a detailed introduction to the various components included in the framework.\par
\textbf{State}: The state is the dialogue context along with the generated response. The state at time $t$, denoted as $s_t=\left( {\hat y_{<t},c} \right)$, comprises the dialogue context $c$ and the tokens generated prior to time $t$.\par
\textbf{Action}: The action is the generated response. The action at time $t$ is denoted as $a_t = \hat y_t$, which corresponds to the token generated at time $t$.\par
\textbf{Policy}: The policy network $\pi_\theta$ generates empathetic responses based on the context, where $\theta$ represents the learnable parameters. At time $t$, the policy is ${\pi_\theta}\left({a_t}|{s_t}\right) = {\pi_\theta}\left({\hat y_t|\hat y_{<t}, c}\right)$, which predicts the current token using the dialogue context and previously generated tokens. EmpRL initializes the policy $\pi_\theta$ using the fine-tuned generator $\rho$.\par 
\textbf{Reward Function}: As shown in Fig.~\ref{fig:emprl}, the reward function consists of two components: empathy reward and KL penalty. The empathy reward aligns the empathy levels between generated and target responses, and KL penalty prevents the policy from deviating far from the generator.\par
(1) \textbf{Empathy Reward}: The emotional reaction reflects affective empathy, while interpretation and exploration represent cognitive empathy. To ensure that empathy levels of the generated responses align more closely with the target responses, we develop an empathy reward function that integrates the three empathy communication mechanisms:
\begin{equation}
\mathcal{R}_{emp}\left(\hat y, y, c\right)=\exp \left( {{\rm{ - }}{{\cal L}_{emp}}\left( {\hat y,y,c} \right)} \right),
    \label{r_emp}
\end{equation}
\begin{equation}
    \mathcal{L}_{emp} \left( {\hat y, y, c} \right)=\sum\limits_i {{\rm{CrossEntropy}}\left( {{{\hat l}_i},{l_i}} \right)},
\end{equation}
\begin{equation}
	{{\hat l}_i} = {\rm{EmpathyIdentifier}}_i \left( {c,\hat y} \right),
\end{equation}
\begin{equation}
	{l_i} = {\rm{EmpathyIdentifier}}_i \left( {c,y} \right).
\end{equation}
where $y$, $\hat y$, and $c$ denote the target response, generated response, and context, respectively; $l_i$ and $\hat {l}_i$ represent the empathy levels of the target response and generated response within the context, respectively, as determined by the pre-trained empathy identifier ${\rm{EmpathyIdentifier}}_i$; and $i$ corresponds to a communication mechanism, which belongs to emotion reaction, interpretation, or exploration.\par
EmpRL treats $l_i$ and $\hat {l}_i$ as the gold and predicted labels, respectively. The cross-entropy loss between them is computed as the loss for communication mechanism $i$. By summing the losses across three different communication mechanisms, we obtain the empathy level loss $\mathcal{L}_{emp} \left( {\hat y, y, c} \right)$. Finally, we use Eq.~(\ref{r_emp}) to product the empathy reward $\mathcal{R}_{emp}\left(\hat y, y, c\right)$.\par
(2) \textbf{KL Penalty}: To prevent the policy from deviating excessively from the fine-tuned generator, which could lead to incoherent responses, a KL-divergence penalty term is incorporated into the empathy reward. The KL penalty at time $t$ is:
\begin{equation}
	\mathcal{R}_{k l}\left(\hat{y}_{<t}, c\right) =\log \frac{\pi\left(. \mid \hat{y}_{<t}, c \right)}{\rho\left(. \mid \hat{y}_{<t}, c\right)}.
\end{equation}
The final reward vector $\mathcal{R}\left(\hat y, y, c\right) \in \mathbb{R}^T$ is defined as:
\begin{equation}
	\mathcal{R}\left(\hat y, y, c\right)=\left\{r_t: t=1, \cdots, T\right\},
\end{equation}
\begin{equation}
    {r_t} = \left\{ \begin{array}{l}
        - \beta {{\cal R}_{kl}}\left( {{{\hat y}_{ < t}},c} \right),\, t = 1, \cdots ,T - 1,\\
        {{\cal R}_{emp}}\left( {\hat y,y,c} \right) - \beta {{\cal R}_{kl}}\left( {{{\hat y}_{ < t}},c} \right),\, t = T.
    \end{array} \right.
\end{equation}
where $r_t$ denotes the reward at time $t$, $T$ represents the length of the generated response, and $\beta$ is the coefficient for the KL penalty.\par
Because the empathy reward $\mathcal{R}_{emp}\left(\hat y, y, c\right)$ is only available once the response is fully generated, it does not contribute to the rewards at any time before $T$.\par
\textbf{Advantage Function}: According to generalized advantage estimator\cite{schulman2016high}, the advantage function for the state-action pair $\left( {{s_t},{a_t}} \right)$ at time $t$ is defined as:
\begin{equation}
	\hat{A}_\pi^t =\delta_t+\left( {\gamma \lambda } \right) \delta_{t+1}+\ldots+\left( {\gamma \lambda } \right)^{T-t+1} \delta_{T-1},
\end{equation}
\begin{equation}
	\delta_t =r_t+\gamma V_\pi\left(\hat{y}_{<t+1}, c\right)-V_\pi\left(\hat{y}_{<t}, c\right).
\end{equation}
where $r_t$ denotes the reward at time $t$, $V_\pi\left(s_t \right)$ represents the value function for state $s_t$, and $\gamma,\lambda \in \left[ {0,1} \right]$ are the discount factor and adjustment factor, respectively.\par
In EmpRL, the policy and value function share a common network architecture. Consequently, an additional trainable token-level value head (V-Head), implemented as a linear layer, is directly added on top of the generator's hidden states to compute the value function.\par
\textbf{Objective}: The objective of EmpRL is to maximize the expected reward of generated responses:
\begin{equation}
	\mathop {\max }\limits_\theta \mathbb{E}_{\hat{y} \sim \pi_\theta}\left[ {{\cal R}\left( {\hat y,y,c} \right)} \right].
\end{equation}

To reduce the variability of predictions, the maximization of the expected reward is reformulated as the maximization of the expected advantage:
\begin{equation}
	\mathop {\max }\limits_\theta \mathbb{E}_{\hat{y} \sim \pi_\theta}\left[ {\sum\limits_{t = 1}^T {\hat A_\pi ^t} \left( {\left( {{{\hat y}_{ < t}},c} \right),{{\hat y}_t}} \right)} \right].
\end{equation}
\textbf{Optimization Method}: We employ the PPO algorithm\cite{schulman2017proximal} to train the policy $\pi_\theta$, and the total loss function is defined as:
\begin{equation}	
    \mathcal{L}_\theta =-\mathcal{L}_\theta^{CLIP}+\alpha \mathcal{L}_\theta^{V F},
\end{equation}
\begin{footnotesize}
\begin{equation}
\!\mathcal{L}_\theta^{CLIP}\!=\!\mathbb{E}_{\hat{y} \sim \pi_\theta}\!\Bigl[\sum_{t=1}^T \min \!\Bigl(c_\pi^t(\theta) \hat{A}_\pi^t,\operatorname{clip}\!\left(c_\pi^t(\theta), 1\!-\!\epsilon, 1\!+\!\epsilon \!\Bigr) \hat{A}_\pi^t\!\right)\!\Bigr],
\end{equation}
\end{footnotesize}
\begin{footnotesize}
\begin{equation}
    \mathcal{L}_\theta^{V F}\!=\!\mathbb{E}_{\hat{y} \sim \pi_\theta}\!\Bigl[\sum_{t=1}^T\Bigl(V_\pi\left(\hat{y}_{<t}, c\right)-\left(\hat{A}_\pi^t+V_{\pi_{\text {old }}}\left(\hat{y}_{<t}, c\right)\right)\Bigr)^2\!\Bigr].
\end{equation}
\end{footnotesize}
where $\mathcal{L}_\theta^{CLIP}$ is the clipped surrogate policy objective function, and $\mathcal{L}_\theta^{V F}$ is the value function squared error term; hyperparameters $\alpha$ and $\epsilon$ represent the weights for the linear combination of losses and the clipping range of proximal policy ratio, respectively; and $c_\pi^t(\theta)=\frac{\pi_\theta\left(a_t \mid s_t\right)}{\pi_{\theta_{old}}\left(a_t \mid s_t\right)}$ denotes the ratio of the new policy to the old policy.\par
Minimizing total loss function involves simultaneously maximizing the clipped surrogate policy objective and minimizing the value error. Maximizing the clipped surrogate policy objective aims to maximize the expected reward, where the surrogate policy objective calculates the ratio between the current and the old policy and then constrains the update step within an appropriate range. This constraint prevents large variations during policy updates, leading to more stable optimization. On the other hand, minimizing the token-level value estimation, which is based on the difference between values of the new policy and the estimated dense returns of the old policy, further enhances the stability.
\section{Experimental Settings}
\label{sec:experiment_settings}
\subsection{Dataset}
\begin{figure}[!t]
    \centering
    \includegraphics[width=3.5in]{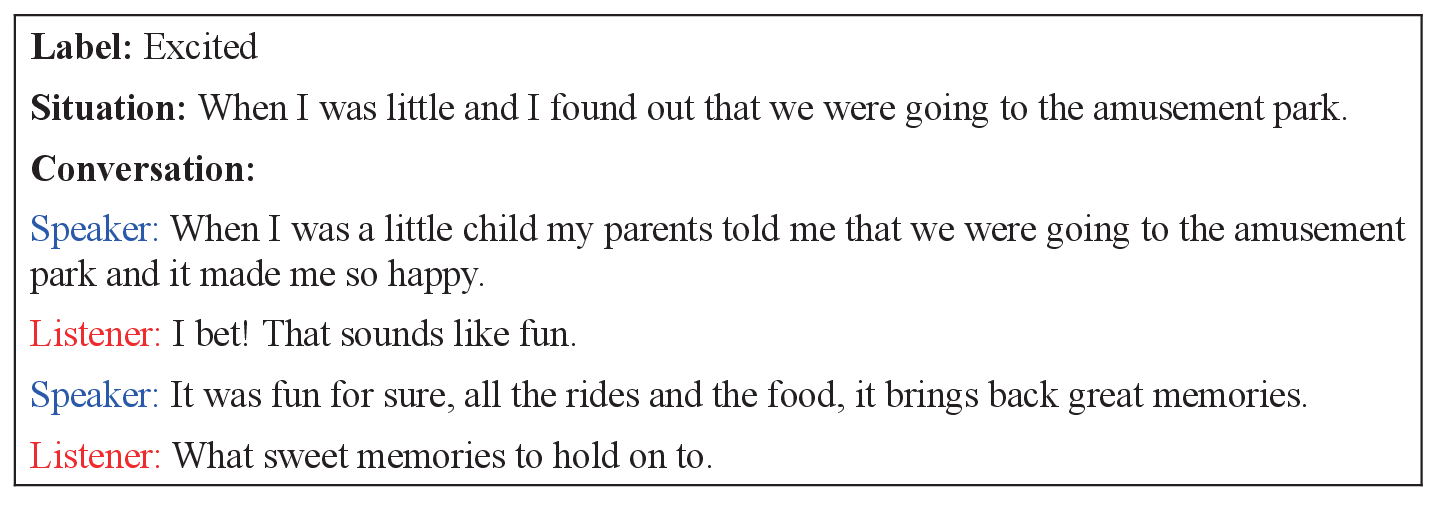}
    \caption{An example from the EmpatheticDialogues dataset.}
    \label{fig:data_examp}
\end{figure}

We conduct experiments on the EmpatheticDialogues dataset\cite{rashkin2019towards}. EmpatheticDialogues\footnote{\url{https://github.com/facebookresearch/EmpatheticDialogues}} is a large-scale multi-turn dialogue dataset containing $24,850$ open-domain empathetic conversations between speakers and listeners. An example is illustrated in Fig.~\ref{fig:data_examp}. Each conversation is grounded in a situation, which is a description provided by the speaker derived from an emotion label (the dataset includes 32 uniformly distributed emotion labels). During the conversation, the speaker talks about the situation, while the listener responds empathetically. Our EmpRL framework generates responses exclusively based on context information, without relying on emotion labels or situation descriptions.

We also use the Persona-based Empathetic Conversation (PEC) dataset\cite{zhong2020towards} to evaluate EmpRL. PEC\footnote{\url{https://github.com/zhongpeixiang/PEC}} is a real-world, persona-grounded empathetic dialogue dataset collected from two subreddits, happy and offmychest on Reddit. Since our task focuses on dyadic conversations and does not incorporate persona information, multi-party conversations are excluded, and persona-related details are disregarded. Furthermore, we ensure that each dialogue context corresponds to a single response. After preprocessing, the dataset contains $56,808$, $7,115$, and $7,131$ samples for training, validation, and testing, respectively. 
\subsection{Implementation Details}
We implement the proposed EmpRL\footnote{Our code is available at \url{https://github.com/butterfliesss/EmpRL}} using Transformers library\cite{wolf-etal-2020-transformers}. The implementation comprises two stages: fine-tuning the generator and further training it using RL.\par
\textbf{Generator Fine-tuning}: We first train the T5-base model using full fine-tuning. During training, we employ the AdamW optimizer\cite{loshchilov2017decoupled}, with an initial learning rate of $1.0e-4$ and the batch size of $8$. During inference, we set the maximum decoding step to $30$ and use the topk-topp\cite{holtzman2019curious} sampling strategy, where $k=20$, $p=1.0$, and temperature $\tau =0.9$.\par
\textbf{RL Training}: We employ the AdamW optimizer during RL training. The number of steps for training the generator using PPO is set to $1,600$, with the learning rate of $1.0e-5$ and the batch size of $32$. The hyperparameters for the KL penalty coefficient $\beta$, discount factor $\gamma$, adjustment factor $\lambda$, linear combination weight between losses $\alpha$, and proximal policy ratio clipping range $\epsilon$ are set to $0.2$, $1.0$, $0.95$, $0.1$, and $0.2$, respectively.
\subsection{Baselines}
We compare the proposed EmpRL with the following empathetic models.\par
\textbf{MoEL}\cite{lin2019moel}: It utilizes different Transformer decoders to compute response representations for each possible user emotion, and then softly combines these representations to generate responses.\par
\textbf{MIME}\cite{majumder2020mime}: A Transformer-based model that takes into account emotion grouping and emotion mimicry. It first generates both mimicking and non-mimicking representations, and subsequently fuses these representations to produce responses.\par
\textbf{EmpDG}\cite{li2020empdg}: A multi-resolution adversarial framework comprising an empathetic generator and interactive discriminators. The generator leverages coarse- and fine-grained emotions to generate responses, while discriminators interact with user feedback.\par
\textbf{KEMP}\cite{li2022knowledge}:
A knowledge-aware model that integrates external commonsense knowledge and emotional lexicon knowledge to enhance its ability to explicitly perceive and express emotions.\par
The models presented above primarily focus on affective empathy, whereas the following baselines consider both affective and cognitive dimensions.\par
\textbf{CEM}\cite{sabour2022cem}: A Transformer-based model that utilizes COMET\cite{bosselut2019comet} to generate various types of commonsense knowledge, thereby
enhancing the understanding of user's situation and feelings.\par
\textbf{SEEK}\cite{wang2022empathetic}: A serial encoding and emotion-knowledge interaction framework that predicts emotion-intent characteristics of responses and models a bi-directional interactive selection process between commonsense knowledge and emotions.\par
\textbf{CASE}\cite{zhou2022case}: It constructs two heterogeneous graphs involving commonsense and conceptual knowledge, and aligns coarse- and fine-grained cognition and affection through a two-level strategy. \par
In addition, we also conduct comparisons with open-source and API-based language models.\par
\textbf{T5}\cite{raffel2020exploring}: We use T5-base as the backbone of our EmpRL framework, which contains 220 million parameters. The pre-trained model is further trained using full fine-tuning.\par
\textbf{Llama3}\cite{touvron2023llama}: We adopt LLama3-8B, an LLM developed by Meta AI, and fine-tune it using LoRA\cite{hulora} to generate responses.\par
\textbf{ChatGPT}\footnote{\url{https://openai.com/blog/chatgpt}}: We utilize the ``gpt-3.5-turbo" model from the OpenAI ChatGPT API and employ a 2-shot prompt setting to generate responses. \par 
%
%
%
%
\subsection{Evaluation Metrics}
We evaluate the performance of EmpRL and baselines using both automatic and human evaluations. Human evaluation consists of human ratings and human A/B tests.\par
\textbf{Automatic Evaluation}: 
We employ the following metrics:
\begin{itemize}
    \item \textbf{Perplexity} (\textbf{PPL}): PPL is the exponent of the cross-entropy and evaluates the overall quality of generated responses. A lower score represents more fluent and natural responses.
    
    \item \textbf{Distinct-1/2 (Dist-1/2)}\cite{li2016diversity}: Dist-1/2 measures the diversity of generated responses by calculating the proportion of distinct unigrams/bigrams. Higher scores indicate greater diversity in the responses.

    \item \textbf{Empathy F1-score (Emp-F1)}: To evaluate the similarity in empathy levels between generated and target responses within a given context, we design the Emp-F1 evaluation metric. Specifically, our pre-trained empathy identifiers are employed to assess the empathy levels for three communication mechanisms in each \texttt{<context, target response>} pair, treating these as gold labels. Meanwhile, empathy levels for the same three communication mechanisms are identified in each \texttt{<context, generated response>} pair and treated as predicted labels. The weighted average F1-scores for each empathy mechanism are then calculated separately. Finally, the average of these three weighted F1-scores yields the Emp-F1 score, which measures the similarity in empathy levels.
\end{itemize}

\textbf{Human Evaluation}: We randomly sample 100 dialogue contexts along with their corresponding responses generated by EmpRL and baseline models. Three graduate students with expertise in natural language processing and affective computing from our research team are recruited as human evaluators. To ensure consistency and reliability, all evaluators receive detailed instructions and complete a brief practice session before assessments. This preparatory phase familiarizes them with the evaluation process and align their interpretations of the evaluation criteria. Human evaluations are conducted under a double-blind setup to mitigate potential biases. The specific instructions for both the human rating and human A/B test are outlined below.\par
\textbf{Human Rating}: The quality of generated responses is assessed based on three criteria: empathy, relevance, and fluency. Ratings for each criterion are on a scale from 1 to 5, where higher scores indicate better performance (1: poor, 2: marginal, 3: moderate, 4: good, 5: excellent).
\begin{itemize}
	\item \textbf{Empathy}: It measures whether the generated responses express an understanding of the user's situation and feelings and convey appropriate emotions.
	\item \textbf{Relevance}: It measures whether the generated responses are relevant to the context.
	\item \textbf{Fluency}: It measures whether the generated responses are natural and fluent.
\end{itemize}
The average score from three evaluators is used as the final rating.\par
\textbf{Human A/B Test}: Given a dialogue context and two candidate responses (A and B), please select the preferred response through a comprehensive assessment of their overall quality. This assessment should take into account multiple dimensions, including empathetic expression, contextual relevance, and linguistic fluency. If both responses are deemed equally effective, a ``Tie'' option is available.\par
The final result is determined by majority vote. In cases where the three evaluators provide conflicting results, a fourth evaluator is introduced. The proportions of ``Win'', ``Lose'', and ``Tie'' outcomes for EmpRL, in comparison to the baselines, are reported.\par
For both human evaluations, we use Fleiss' Kappa ($\kappa $)\cite{fleiss1973equivalence} to measure the inter-evaluator agreement.
\section{Results and Analysis}
\label{sec:results}
\subsection{Automatic Evaluation Results}
The automatic evaluation results on EmpatheticDialogues are presented in Table~\ref{tab:auto_results}. We conduct EmpRL with $5$ random seeds and report the mean scores along with $95\%$ confidence intervals. From the table, we conclude that: (1) Models such as CASE, SEEK, and CEM, which incorporate both affective and cognitive empathy, outperform models that focus solely on affective empathy, such as MoEL, MIME, and KEMP. This validates that considering both types of empathy leads to better response generation. (2) In terms of PPL, EmpRL shows a reduction of $22.69$ compared to the state-of-the-art baseline CASE, indicating that leveraging pre-trained language models can generate more fluent responses. Llama3 achieves a lower PPL primarily because it is an LLM with more extensive training datasets and parameters. (3) Regarding Dist-1/2, EmpRL performs comparably to Llama3 and ChatGPT, demonstrating that the proposed EmpRL is capable of generating diverse responses. (4) In terms of Emp-F1, EmpRL achieves the highest score of $69.43\%$. This result indicates that its responses exhibit a greater similarity in empathy levels to the target responses, further validating the effectiveness of EmpRL.
\begin{table}[!t]
    \centering
    \caption{Automatic evaluation results on EmpatheticDialogues, with confidence intervals reported. $\uparrow$ means a higher score is better whereas $\downarrow$ is exactly the opposite. Numbers in bold denote the best results. $\dagger$, $\ddagger$ represent that the improvement over the best and second-best baselines is statistically significant (one-sample t-test, $p$-value $<0.05$), respectively.}
    \renewcommand\arraystretch{1.25}
    \begin{tabular}{lcccc}
        \toprule
        Models          & PPL$\downarrow$    & Dist-1$\uparrow$ & Dist-2$\uparrow$ & Emp-F1$\uparrow$ \\ 
        \midrule
        MoEL            & 36.97 & 0.44       & 2.16       & 59.31 \\
        MIME            & 37.36  & 0.47       & 2.02  & 59.69      \\
        EmpDG           &     37.26   &     0.49       &     2.23  & 58.54     \\
        KEMP            & 36.39  & 0.61       & 2.65      &  54.29 \\ \hline
        CEM             & 36.12  & 0.66       & 2.99     & 63.32  \\
        SEEK      &    37.25    &     0.72       &    3.11   & 63.08 \\
        CASE            & 35.59  & 0.78       & 3.98    & 64.72   \\ \hline
        T5 & 11.95 & 4.01 & 25.24 & 65.70 \\
        Llama3 & \textbf{7.78} & \textbf{4.77} & 28.78 & 67.74 \\
        ChatGPT         &    -    &    4.22        &   23.99    & 64.79  \\ \hline
        EmpRL   &    12.90 & 4.65$^\ddagger$ & \textbf{29.03}$^\dagger$ & \textbf{69.43}$^\dagger$      \\
        & $\pm$0.10 &  $\pm$0.07 &  $\pm$0.15 & $\pm$0.19\\ 
        \bottomrule
    \end{tabular}
    \label{tab:auto_results}
\end{table}

We also conduct experiments on the PEC dataset\footnote{Since PEC does not contain emotion labels, models such as KEMP, CEM, and CASE cannot be used as baselines.}. As shown in Table~\ref{tab:pec_auto_results}, we obtain similar results to those on the EmpatheticDialogues dataset, with Dist-1/2 scores comparable to Llama3 and ChatGPT, and the best score in terms of Emp-F1. The results demonstrate the robustness of our proposed EmpRL framework.
\begin{table}[!t]
	\centering
	\caption{Automatic evaluation results on PEC, with confidence intervals reported. $\dagger$, $\ddagger$ represent that the improvement over the best and second-best baselines is statistically significant (one-sample t-test, $p$-value $<0.05$), respectively.}
	\renewcommand\arraystretch{1.25}
	\begin{tabular}{lcccc}
		\toprule
		Models          & PPL$\downarrow$    & Dist-1$\uparrow$ & Dist-2$\uparrow$ & Emp-F1$\uparrow$ \\ 
		\midrule
		T5 & 18.89 & 4.30  &  25.51 &  69.84 \\
		Llama3 & \textbf{13.78} & \textbf{5.32} & \textbf{28.79} & 70.98 \\
		ChatGPT         &    -    &    4.39        &   25.10    & 67.44  \\ \hline
		EmpRL   &  20.25 &   4.96$^\ddagger$ & 28.43$^\ddagger$ & \textbf{72.73}$^\dagger$      \\ 
		& $\pm$0.18 &  $\pm$0.20 &  $\pm$0.45 & $\pm$0.18\\ 
		\bottomrule
	\end{tabular}
	\label{tab:pec_auto_results}
\end{table}
\subsection{Human Evaluation Results}
\textbf{Human Rating Results}: Table~\ref{tab:human_rating} presents human rating results for EmpRL and baselines. First, CASE achieves the highest performance among the task-related baselines, demonstrating the effectiveness of incorporating both affective and cognitive empathy. Second, EmpRL significantly outperforms the task-related baselines across all aspects, indicating that the proposed framework effectively enhances the quality of generated responses. Third, EmpRL surpasses Llama3 in terms of empathy, while both models achieve comparable results in relevance and fluency, further validating EmpRL's capability to generate more empathetic responses. Finally, ChatGPT outperforms EmpRL, primarily for the following reasons: ChatGPT benefits from more diverse training datasets, which enable it to possess a broader knowledge base; it is initialized with GPT-3.5, which has a much large number of parameters; and it leverages reinforcement learning from human feedback, which aids in aligning with complex human values. The Fleiss' Kappa scores are above $0.2$ across all models, indicating fair agreement ($\kappa  \in \left( {0.2,\;0.4} \right]$) among the evaluators. Due to the subjective nature of the evaluation, where different evaluators have varying criteria for giving scores from 1 to 5, the results of inter-evaluator agreement are not particularly high. Nonetheless, the scores still demonstrate a certain level of agreement.\par
\begin{table}[!t]
	\centering
	\caption{Human rating results on EmpatheticDialogues. $\ddagger$ represents that the improvement over the second-best baseline is statistically significant (independent t-test, $p$-value $<0.01$), and $95\%$ confidence intervals are reported.}
	\renewcommand\arraystretch{1.25}
	\begin{tabular}{lcccc}
		\toprule
		Models          & Empathy   & Relevance & Fluency & Kappa \\ 
		\midrule
		MoEL            & 2.11 & 2.32       &  3.44  & 0.27   \\
		CEM & 2.22 & 2.43 & 3.61 & 0.28\\
		CASE      & 2.27&  2.48   &  3.65  & 0.27    \\ 
		Llama3 & 3.20 & 3.47 & 4.06 & 0.31 \\
		ChatGPT    &   \textbf{4.19}   &   \textbf{4.39}       &    \textbf{4.61}  & 0.24    \\
		\hline
		EmpRL        &   3.73$^\ddagger$     &     3.79$^\ddagger$       &     4.21$^\ddagger$   & 0.27      \\ 
		& $\pm$0.10 &  $\pm$0.10 &  $\pm$0.07 & \\
		\bottomrule
	\end{tabular}
	\label{tab:human_rating}
\end{table}
\begin{table}[!t]
	\centering
	\caption{Human A/B test results on EmpatheticDialogues. $\dagger$, $\ddagger$ represent significant improvement with $p$-value $<0.01/0.05$ (sign test), respectively.}
	\renewcommand\arraystretch{1.25}
	\begin{tabular}{lcccc}
		\toprule
		Models          & Win   & Lose  & Tie & Kappa \\ 
		\midrule
		EmpRL vs MoEL            & \textbf{72.0}$^\dagger$ &   12.0   &  16.0 & 0.44\\
		EmpRL vs CEM & \textbf{73.0}$^\dagger$ & 14.0 & 13.0 & 0.45 \\
		EmpRL vs CASE            &  \textbf{68.0}$^\dagger$ &  17.0  &  15.0 & 0.47\\ 
		EmpRL vs Llama3            & \textbf{42.0}$^\ddagger$ &   35.0   &  23.0 & 0.43\\
		EmpRL vs ChatGPT         &  10.0 &    44.0 &   \textbf{46.0}  & 0.41 \\
		\bottomrule
	\end{tabular}
	\label{tab:human_ab}
\end{table}
\textbf{Human A/B Test Results}: The results of human A/B test are shown in Table~\ref{tab:human_ab}. Consistent with the human ratings, EmpRL outperforms task-related baselines and Llama3, demonstrating the effectiveness of the proposed RL-based framework. Furthermore, EmpRL achieves a ``Win'' or ``Tie'' ratio of $54.0\%$ compared to ChatGPT. This finding indicates that although EmpRL does not surpass ChatGPT, it remains competitive in the majority of generated responses, further demonstrating the superiority of EmpRL. It is worth noting that all Kappa scores exceed $0.4$, indicating moderate agreement ($\kappa  \in \left( {0.4,\;0.6} \right]$) among the evaluators, which reflects a degree of reliability in the evaluations. However, notable discrepancies remain, primarily due to the inherent complexity of the task and the subjectivity of human judgment. Empathy is a complex and multifaceted concept, and both understanding and expressing it pose significant challenges, not only for dialogue systems but also for humans. Thus, addressing the task of empathetic response generation, with a focus on empathy factors, as well as developing evaluation methods that ensure consistency and reliability, requires further exploration.\par
\begin{figure}[!t]
	\centering
	\hspace{-20mm}
	\subfloat[ChatGPT]{\label{fig:Chatgpt}
		\includegraphics[width=2.6in]{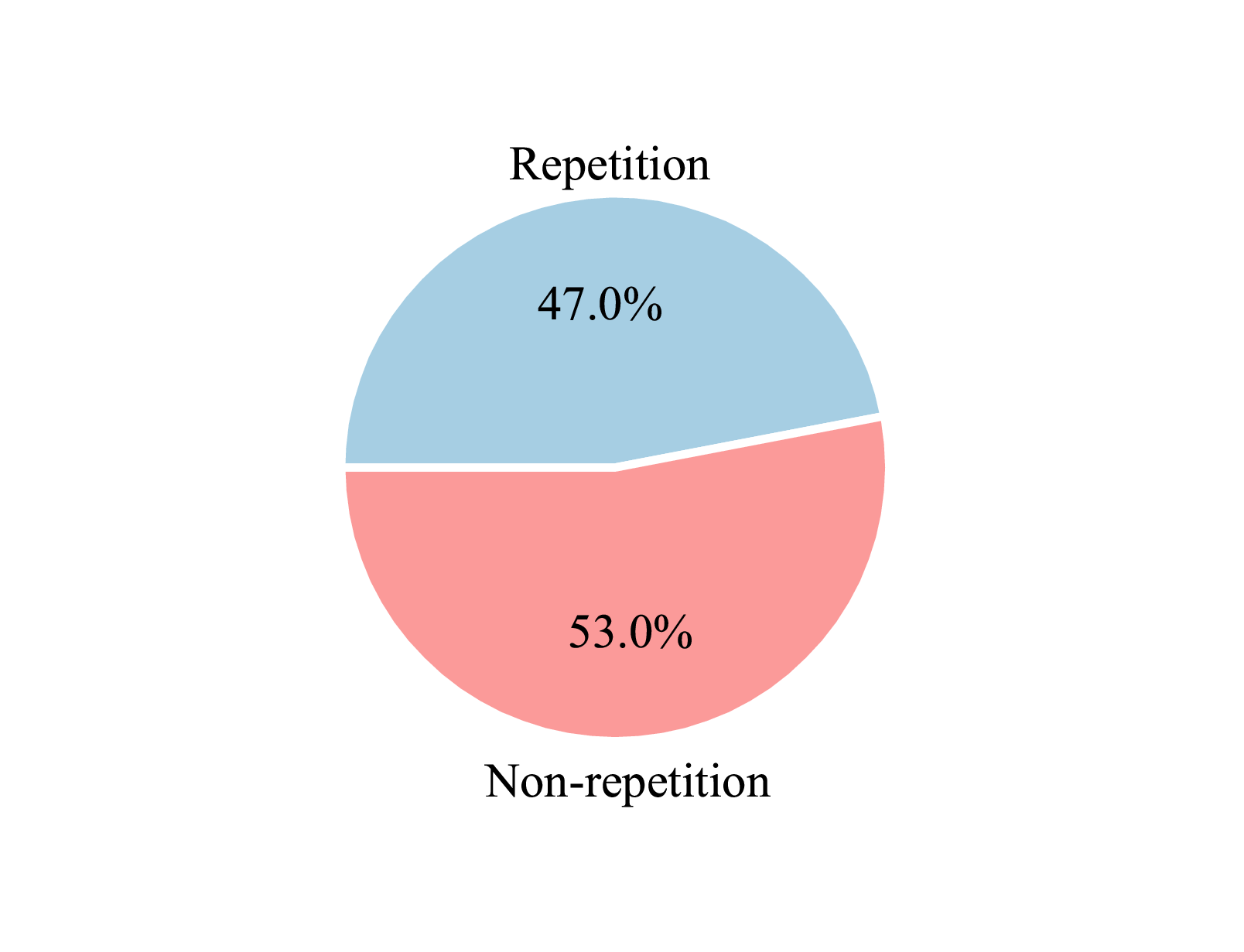}}
	\hspace{-30mm}
	\subfloat[EmpRL]{\label{fig:Emp}
		\includegraphics[width=2.6in]{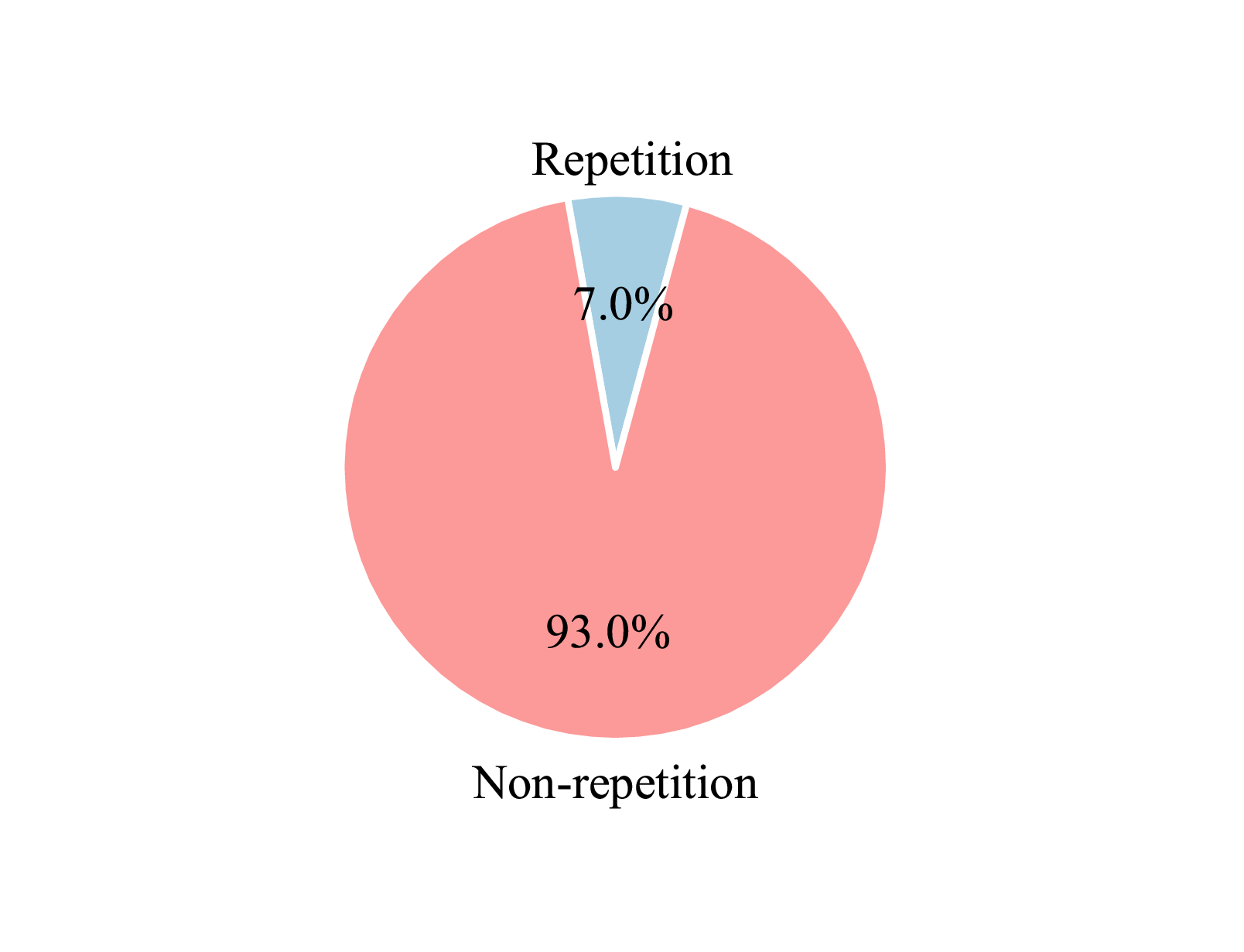}}
	\hspace{-15mm}
	\caption{Proportions of repetitive and non-repetitive patterns across models on EmpatheticDialogues.}
	\label{fig:repetitive_patterns}
\end{figure}

On the other hand, as observed by Zhao et al.\cite{zhao2023chatgpt}, ChatGPT often repeats the pattern of ``emotional restatement + information expansion'' when expressing empathy. To further validate this observation, we perform a manual inspection of the responses generated by ChatGPT and EmpRL for human evaluation samples, assessing whether their responses exhibit this repetitive pattern. The results, including the repetition proportion (defined as the percentage of responses containing repetitive patterns) and the non-repetition proportion, are presented in Fig.~\ref{fig:repetitive_patterns}. We find that the proportion of ChatGPT repeating this pattern is as high as $47.0\%$. This repetitive pattern may cause users to experience monotony and a lack of freshness. Therefore, enhancing personalized empathy capabilities requires greater attention.\par 
\subsection{Ablation Study}
To investigate the effectiveness of different components in our EmpRL framework, we conduct ablation experiments. We consider the following three settings:
\begin{itemize}
    \item \textbf{w/o KL Penalty}: Remove the KL penalty term from the reward function.
    \item \textbf{w/o Empathy Reward}: Remove the empathy reward from the reward function.
    \item \textbf{w/o RL}: Remove RL training and use only the fine-tuned generator to generate responses.
\end{itemize}
\begin{table}[!t]
	\centering
	\caption{Ablation study on EmpatheticDialogues.}
	\renewcommand\arraystretch{1.25}
	\begin{tabular}{lcccc}
		\toprule
		Models&
		PPL$\downarrow$    & Dist-1$\uparrow$ & Dist-2$\uparrow$ & Emp-F1$\uparrow$  \\  
		\midrule
		EmpRL           &    12.90 & 4.65 & 29.03 & \textbf{69.43}         \\ 
		w/o KL Penalty & 14.49 & \textbf{4.79} & \textbf{29.88} & 68.87 \\
		w/o Empathy Reward & 12.10 & 3.96 & 25.13 & 65.90 \\
		w/o RL & \textbf{11.95} & 4.01 & 25.24 & 65.70 \\ 
		\bottomrule
	\end{tabular}
	\label{tab:ablation}
\end{table}
The ablation results\footnote{Removing the KL penalty term leads to a significant increase in PPL as the number of training steps grows. Therefore, the result of ``w/o KL Penalty'' is obtained by fine-tuning the generator with PPO for 600 steps.} are reported in Table~\ref{tab:ablation}. From the table, we observe that: (1) Removing the KL penalty term results in a significant decrease in the fluency of generated responses. Hence, integrating the KL penalty into the reward function is necessary. (2) Removing the empathy reward causes a notable  reduction in Dist1/2 and Emp-F1 scores. This indicates that the designed empathy reward function improves the diversity of generated responses and aligns their empathy levels more closely with target responses. (3) Without RL training, the performance is similar to that observed when the empathy reward is removed---there is a moderate decrease in PPL, while Dist2 and Emp-F1 exhibit more significant drops. In summary, the ablation results demonstrate that introducing RL for further training the generator, designing an empathy reward, and integrating the KL penalty term in the reward function are all beneficial for enhancing the quality of generated empathetic responses.
\subsection{Discussion}
\begin{figure}[!t]
	\centering
	\includegraphics[width=2.6in]{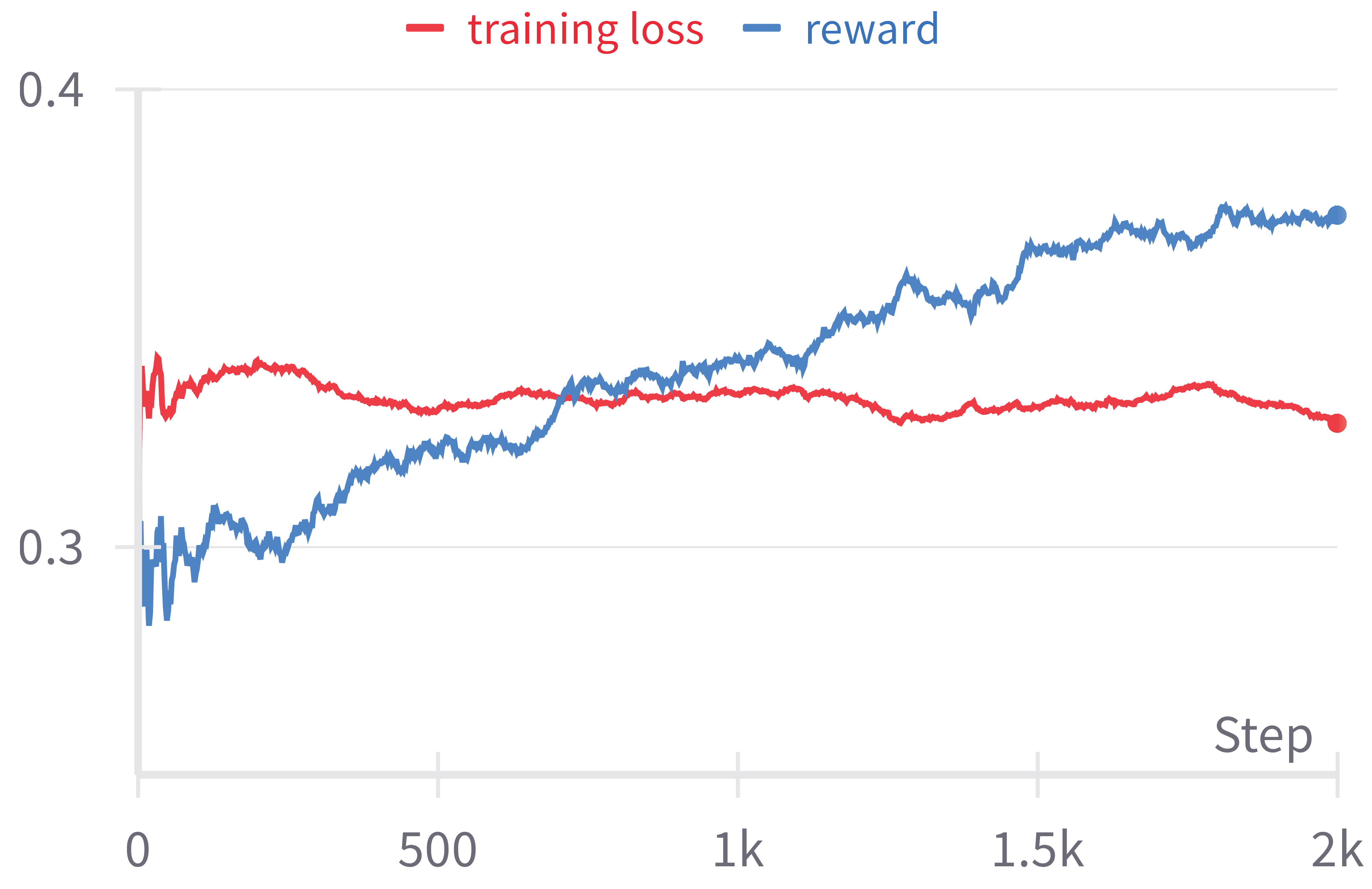}
	\caption{Trends in the total training loss and reward score during RL training on EmpatheticDialogues.}
	\label{fig:train_loss}
\end{figure}

\textbf{Analysis on RL Training}: During the RL training, we present the trends of the total training loss and reward score on the EmpatheticDialogues dataset to assess the performance of the training process. The trends\footnote{The smoothed results are illustrated.} are depicted in Fig.~\ref{fig:train_loss}. As training progresses, we observe a gradual decrease in the total loss, while the reward score shows a more significant increase. Both metrics stabilize after approximately $1,600$ steps. This indicates the successful training and convergence of the RL process, thereby
demonstrating the effectiveness of our proposed EmpRL framework.

\textbf{Analysis on Open-domain Dialogue Performance}: To explore the performance of EmpRL in open-domain dialogues, we carry out experiments on the DailyDialog dataset\cite{li2017dailydialog}. DailyDialog\footnote{\url{http://yanran.li/dailydialog}} is a high-quality chit-chat dataset containing human conversations about daily life. We assess ChatGPT and our trained models (i.e., T5, Llama3, and EmpRL) on the DailyDialog test set. Additionally, we present the results of several models reported in CTRLStruct\cite{yin2023ctrlstruct} for comparison. BLEU-1/2\cite{papineni2002bleu} and ROUGE-L\cite{lin2004rouge} are used to evaluate model performance.

Table~\ref{tab:daily_auto_results} presents the experimental results. First, it is evident that EmpRL shows some gaps when compared to Llama3 and ChatGPT, which is understandable given their advanced natural language generation capabilities. Second, EmpRL outperforms Seq2Seq, BART, and CTRLStruct in terms of BLEU-2 and ROUGE-L, validating the effectiveness of our model in open-domain dialogues. Finally, EmpRL achieves results comparable to T5, indicating that EmpRL is not overly fine-tuned for empathy alone but still retains certain general response generation capabilities.
\begin{table}[!t]
	\centering
	\caption{Experimental results on DailyDialog.}
	\renewcommand\arraystretch{1.25}
	\begin{tabular}{lcccc}
		\toprule
		Models          & BLEU-1$\uparrow$    & BLEU-2$\uparrow$ & ROUGE-L$\uparrow$  \\ 
		\midrule
		Seq2Seq            & 30.90 & 3.60       & 4.90    \\
		BART            & 36.40  & 14.10       & 7.50 \\
		CTRLStruct & 39.70 & 15.70 & 8.00 \\
		\hline
		T5             & 30.46  & 18.66       & 9.64   \\
		Llama3            & 31.49  & 19.78       & 12.28   \\ 
		ChatGPT         &    39.39 &  24.59  &    12.84        \\ \hline
		EmpRL      &    28.76   &     17.66       &    9.21   \\
		\bottomrule
	\end{tabular}
	\label{tab:daily_auto_results}
\end{table}

\subsection{Case Study}
\begin{table*}[!t]
	\centering 
	\caption{Case study on EmpatheticDialogues.}
	\renewcommand\arraystretch{1.25}
	\resizebox{\textwidth}{!}{
		\begin{tabular}{ll}
			\toprule
			\textbf{Case 1} & \\ 
			Context & \textit{Speaker:} I received a phone call the other day from a friend I hadn't heard from in ages.\\ \hline
			MoEL & That is awesome. Did you get a lot?\\
			CEM & That is nice of them.\\
			CASE & Oh wow, that is a lot of money!\\
			Llama3 & How did that go?\\
			ChatGPT & That's always a nice surprise. What did you guys talk about?\\
			EmpRL & That's wonderful. It's always good to be able to reconnect to family and friends.\\
			Reference & Oh wow, I bet that was nice. Hopefully y'all had a good talk.\\ \hline \hline
			\textbf{Case 2} & \\ 
			Context & \textit{Speaker:} A lot people know Toys-R-Us has completely shut down not long ago.\\ \hline
			MoEL & That is a bummer. I am sorry to hear that.\\
			CEM & I am sorry to hear that.\\
			CASE & Oh no! What happened?\\
			Llama3 &  I think I heard about that. I think it was because they weren't as good at selling merchandise online.\\
			ChatGPT & Yes, it's sad to see such a well-known brand go. I hope the employees were able to find new jobs. \\
			EmpRL & Ouch! You must be really sorry. Is there another place you can take the kids to?\\
			Reference & Yea I remember it seeing go away a few months back, sad.\\ \hline \hline
			\textbf{Case 3} & \\ 
			\multirow{5}{*}{Context} & \textit{Speaker:} The other day I had to take my dogs out before work, but my neighbor was right outside  with her aggressive \\
			~ & yappy dog smoking so I couldn't get out.\\
			~& \textit{Listener:} Eww, I hate neighbors that have no class.\\
			~& \textit{Speaker:} She usually cares enough to at least time her cigarettes so they don't get in the way with things like work, \\
			~ & but not this time.\\ \hline 
			MoEL &  I would be too. I would be too.\\
			CEM & I would have been too.\\
			CASE &  I am sorry to hear that.\\
			Llama3 & I would have to do something about that, I'm not a fan of dogs either.\\
			ChatGPT & That's frustrating. Maybe you could talk to her about it and find a solution that works for both of you.\\
			EmpRL &  I see. I know what you mean. We all have trouble with our neighbors.\\
			Reference &  Maybe you should start a petition in the neighborhood to kick her out.\\
			\bottomrule
	\end{tabular}}
	\label{tab:case}
\end{table*}
Table~\ref{tab:case} shows several responses generated by EmpRL and baseline models. In Case $1$, task-related baselines, namely MoEL, CEM, and CASE, demonstrate a limited understanding of the speaker's utterance, resulting in responses that lack contextual relevance. Meanwhile, the response generated by Llama3 is deficient in empathy. In contrast, both ChatGPT and EmpRL effectively comprehend the speaker's situation and feelings, generating empathetic responses. Notably, EmpRL's response expresses empathy in both affective (``That's wonderful.'') and cognitive (``It's always good to be able to reconnect to family and friends.'') aspects. In Case $2$ and $3$, MoEL, CEM, and CASE again fail to generate high-quality empathetic responses, whereas the performance of Llama3, ChatGPT, and EmpRL is significantly superior.\par
From these examples, it can be seen that while ChatGPT is capable of generating high-quality responses,  it typically follows a pattern of ``emotional restatement + information expansion'' to express empathy. For instance,  ChatGPT responds with  ``Yes, it's sad to see such a well-known brand go.'' followed by ``I hope the employees were able to find new jobs.'' in Case $2$. In contrast, our proposed EmpRL framework directly understands the speaker's situation and feelings, generating an empathetic response that includes both affective (``Ouch! You must be really sorry.'') and cognitive (``Is there another place you can take the kids to?'') dimensions.
\section{Conclusion}
\label{sec:conclustion}
In this paper, we propose an EmpRL framework for empathetic response generation. The framework defines an empathy reward function and maximizes the expected reward using the PPO algorithm to align the empathy levels between generated and target responses within a given context. The empathy reward function, which incorporates emotional reaction, interpretation, and exploration communication mechanisms, is constructed utilizing pre-designed and pre-trained empathy identifiers. This reward aligns the empathy levels of generated responses more closely with target responses. For model evaluation, we introduce an Emp-F1 metric to measure the similarity in empathy levels between generated and target responses. Automatic and human evaluations demonstrate that EmpRL enhances the quality of generated responses, producing more empathetic responses that encompass both affective and cognitive dimensions. 
\section{Limitations and Future Work}
\label{sec:futurework}  
The main limitation of our work is that we only validate the effectiveness of aligning the empathy levels between responses generated by T5 and human responses. Our framework, however, is also applicable to other language models. In future work, we will extend this framework to LLMs with the goal of enhancing their empathetic capabilities. Besides, our designed reward function is a single-turn reward, which does not take into account empathy consistency across multi-turn dialogues. In the future, we plan to develop a multi-turn (i.e., trajectory-level) reward and adopt multi-turn RL that not only aligns empathy levels at the individual turn level but also penalizes sudden shifts in empathy, encouraging empathy consistency throughout the dialogue. Furthermore, it would be promising to integrate the retrieval-augmented generation method into our framework. We will create an empathetic dialogue retrieval database to store context-response pairs along with their corresponding empathy levels, and design an efficient retrieval mechanism to further enhance the generation of empathetic responses by retrieving relevant dialogue samples.
\section*{Acknowledgments}
This work is partially supported by Anhui Provincial Natural Science Foundation (2408085QF188) and Natural Science Foundation of China (62376051, 62376084).


 
%

\bibliographystyle{IEEEtran}
\bibliography{myreference.bib}

\end{document}